\documentclass[10pt,journal,compsoc]{IEEEtran}

\usepackage{epsfig}
\usepackage{graphicx}
\usepackage{amsmath}
\usepackage{amssymb}
\usepackage{amsthm} 
\usepackage{algorithm2e}
\usepackage{enumitem}
\usepackage{multirow}
\usepackage{array}
\usepackage{cite}
\usepackage{subfig}
\usepackage{color}
\usepackage{eso-pic}
\usepackage{url}
\usepackage{xspace}

\ifCLASSINFOpdf

\else
\fi

\newcommand{\approximate}{\raise.17ex\hbox{$\scriptstyle\sim$}}

\newcommand{\NTnote}[1]{\textcolor{black}{{ #1}}}

\begin{document}
\title{Coreset-Based Adaptive Tracking}
\author{Abhimanyu Dubey, Nikhil Naik, Dan Raviv, Rahul Sukthankar, and Ramesh Raskar
\IEEEcompsocitemizethanks{\IEEEcompsocthanksitem Abhimanyu Dubey is an undergraduate student at the Indian Institute of Technology, New Delhi, India.
\IEEEcompsocthanksitem Rahul Sukthankar is a research scientist at Google.
\IEEEcompsocthanksitem Nikhil Naik, Dan Raviv, and Ramesh Raskar are with 
the Media Laboratory at Massachusetts Institute of Technology.}
\thanks{}}

\markboth{}%
{Shell \MakeLowercase{\textit{et al.}}: Bare Demo of IEEEtran.cls for Computer Society Journals}
\captionsetup{belowskip=0pt,aboveskip=4pt}
\captionsetup{font=small}
\IEEEtitleabstractindextext{%
\begin{abstract}
We propose a method for learning from streaming visual
data using a compact, constant size representation of all
the data that was seen until a given moment. Specifically,
we construct a ``coreset'' representation of streaming data
using a parallelized algorithm, which is an approximation of 
a set with relation to the squared
distances between this set and all other points in its ambient
space. We learn an adaptive object appearance model from the coreset
tree in constant time and logarithmic space and use it for object
tracking by detection. Our method obtains excellent results for object
tracking on three standard datasets over more than 100 videos. The
ability to summarize data efficiently makes our method ideally suited
for tracking in long videos in presence of space and time
constraints. We demonstrate this ability by outperforming a variety of
algorithms on the TLD dataset with 2685 frames on average. This
coreset based learning approach can be applied for both real-time
learning of small, varied data and fast learning of big data.
\end{abstract}

}

\maketitle
\IEEEdisplaynontitleabstractindextext
\IEEEpeerreviewmaketitle

\IEEEraisesectionheading{\section{Introduction}\label{sec:introduction}}
Real-world computer vision systems are increasingly required to 
provide robust real-time performance under memory and bandwidth 
constraints for tasks such as tracking, on-the-fly 
object detection, and visual search. Robots with visual sensors may be 
required to perform all of these tasks while interacting with 
their environment. On another extreme, IP cameras, 
life logging devices, and other visual sensors may want to hold a 
summary of the data stream seen over a long time.
Hence, new algorithms that enable real-world 
computer vision applications under a tight space-time budget are a necessity. 
In this paper, we propose new algorithms with low space-time complexity 
for learning from streaming visual data, inspired by research in computational 
geometry. 
 
In recent years, computational geometry researchers have proposed 
techniques to obtain small representative sets of 
points from larger sets with constant or 
logarithmic space-time complexity~\cite{badoiu2002approximate,
feldman2013turning,RelativeErrors:Ghashami:14}. 
We propose to adapt these techniques to retain a summary of visual data and train 
object appearance models in constant time and space from this 
summary. Specifically, we perform summarization 
of a video stream based on ``coresets'', which are defined 
as a small set of points that approximately 
maintain the properties of the original set 
with respect to the distance between the set and 
other structures in its ambient space with theoretical guarantees~\cite{HarPeled2004}.
We demonstrate that the coreset formulation is useful for 
learning object appearance models for detection and tracking. 
Coresets have been primarily used as a technique to convert ``big data'' 
into manageable sizes~\cite{KSegmentation:Rosman:14}. 
In contrast, we employ coresets for reducing small data to ``tiny data'', 
and perform real-time classifier training. 

We use the object appearance model learned from the coreset 
representation,  for visual tracking by detection. Tracking systems typically retain 
just an appearance model for the object~\cite{zhong2012scm,jepson2003robust} 
and optionally, data from some prior frames~\cite{supancic2013self,gao2014tgpr}. 
In contrast, we compute a summary of all the previous frames in 
constant time (on average) and update our appearance model continuously with 
this summary. The summarization process improves the 
appearance model with time. As a result, our method is beneficial for 
tracking objects in very long video sequences that need to deal with 
variation in lighting, background, and object appearance, along with 
the object leaving and re-entering the video.

We demonstrate the potential of this technique with 
a simple tracker that combines detections from a linear SVM trained from 
the coreset with a Kalman filter. We call this no-frills 
tracker the ``Coreset Adaptive Tracker (CAT).'' CAT obtains 
competitive performance on three standard datasets, 
the CVPR2013 Tracking Benchmark~\cite{wu2013online}, 
the Princeton Tracking Benchmark~\cite{song2013tracking}, and
the TLD Dataset~\cite{kalal2012tracking}. Our method is 
found to be superior to popular methods 
for tracking longer video sequences from the TLD dataset. 

We must emphasize that the primary goal of this paper
is not to develop the best algorithm for tracking---which is 
evident from our simple design. Rather, our goal is 
to demonstrate the potential of coresets as a data
reduction technique for learning from visual data.
We summarize our contributions below.

\subsection{Contributions} 
\begin{itemize}[leftmargin=*]
\item We propose an algorithm for learning and detection of  
objects from streaming data by training a classifier  
in constant time using coresets. We construct coresets using a parallel
algorithm and prove that this algorithm requires constant time (on average) and 
logarithmic space. 

\item We introduce the concept of  
hierarchical sampling from coresets, which improves learning performance 
significantly when compared to the original formulation~\cite{feldman2013turning}.

\item We obtain competitive results on standard 
benchmarks~\cite{wu2013online,song2013tracking,kalal2012tracking}.
Our method is especially suited for longer videos (with several hundred frames), 
since it summarizes data from \textit{all the frames} in the video so far. 
\end{itemize}

\section{Related Work}
First, we introduce the literature on coresets, 
followed by a summary of the 
prior art on online learning for tracking by detection.\\\\
\textbf{Coresets and Data Reduction:}
The term \emph{coreset} was first introduced by 
Agarwal et al.~\cite{agarwal2004approximating} 
in their seminal work on k-median and k-means problems. 
Badoiu et al.~\cite{badoiu2002approximate} propose a coreset for 
clustering using a subset of input points to generate the
solution, while Har-Peled and Mazumdar~\cite{HarPeled2004} 
provide a different solution for coreset construction 
which includes points that may not be a subset of the original set. 
Feldman et al.~\cite{feldman2007ptas} demonstrate that 
a weak coreset can be generated independent of number of 
points in the data or their feature dimensions. We follow the formulation 
of Feldman et al.~\cite{feldman2013turning}, which 
provides an eminently parallelizable streaming algorithm for 
coreset construction with a theoretical error bound. 
This formulation has been recently used for applications 
in robotics~\cite{Krobots:feldman:13} and for
summarization of time series data~\cite{KSegmentation:Rosman:14}. \\\\
\textbf{Online Learning for Detection and Tracking:}
The problem of object tracking by detection has been 
well studied in computer vision~\cite{SmeuldersTPAMI2014}.
Early important examples of learning based methods include
~\cite{black1996eigentracking,comaniciu2003kernel,jepson2003robust}. 
Ross et al.~\cite{ross2008incremental} learn a low dimensional 
subspace representation of object 
appearance using incremental algorithms for PCA. 
Li et al.~\cite{li2011real} propose a compressive sensing 
tracking method using efficient $l_1$ norm minimization. 
Discriminative algorithms solve the tracking problem 
using a binary classification model for object appearance. 
Avidan~\cite{avidan2004support} combines optical 
flow with SVMs and performs tracking by minimizing the SVM 
classification score, and extends this method with  
an ``ensemble tracker'' using Adaboost~\cite{avidan2007ensemble}.
Grabner et al.~\cite{grabner2008semi} develop 
a semi-supervised boosting framework to tackle the problem of drift 
due to incorrect detections. 
Kwon et al.~\cite{kwon2010visual} represent the object 
using multiple appearance models, obtained from sparse 
PCA, which are integrated using an MCMC framework.
Babenko et al.~\cite{babenko2011robust} introduce the 
semi-supervised multiple instance learning (MIL) algorithm. 
Hare et al.~\cite{hare2011struck} learn an online structured 
output SVM classifier for adaptive tracking. 
Zhang et al.~\cite{zhang2012real} 
develop an object appearance model based on efficient feature 
extraction in the compressed domain and use the features 
discriminatively to separate the object from its background. Kalal 
et al.~\cite{kalal2012tracking} introduce the TLD algorithm, 
where a pair of ``expert'' learners select 
positive and negative examples in a semi-supervised 
manner for real-time tracking. 
Gao et al.~\cite{gao2014tgpr} combine results of a tracker
learnt using  prior knowledge from auxiliary samples and 
another tracker learnt with samples from recent frames.  
Our work contributes to this literature with a novel coreset
representation for model-based tracking.
  
\section{Constructing Coresets}
\label{sec:theory}
\newtheorem{Thm}{Theorem}
In this section, we introduce the key theoretical ideas behind 
coresets and low-rank approximation and provide new results for 
the time and space complexity of a parallel
algorithm for coreset construction. 

\subsection{Coresets} 
We denote time series data of $d$ dimensions as rows in a matrix $A$,
and search for a subset or a subspace of $A$ such that 
the sum of squared distances from the rest of the domain to 
this summarized structure is an $(1+\epsilon)$-approximation of that of 
$A$, where $\epsilon$ is a small constant ($0\leq\epsilon\leq1$).
Specifically, we call $\tilde A$ a `coreset' of $A$ if
\begin{eqnarray}
\label{eq:def1}
 (1-\epsilon) \text{dist}^2(A,S) &\le& \text{dist}^2(\tilde A,S) + c  \nonumber \\
 &\le& (1+\epsilon) \text{dist}^2(A,S),
\end{eqnarray}
where $\text{dist}^2(A,S)$ denotes the sum of squared 
distances from each row on $A$ to its closest 
point in $S$ and $c$ is constant.
Eq.~(\ref{eq:def1}) can be rewritten in a matrix form using 
the Frobenius norm~\cite{feldman2013turning} as
\begin{eqnarray}
 (1-\epsilon) \Vert AY \Vert_F^2 \le \Vert \tilde A Y \Vert_F^2 +c \le (1+\epsilon) \Vert AY \Vert_F^2,
\end{eqnarray}
where $Y$ is a $d\times (d-k)$ orthogonal matrix.
Going forward we denote $\tilde A \approx A$ 
if $\tilde A$ is a $(1+\epsilon)$-approximation of $A$. 
\NTnote{
The best low dimensional approximation can be 
obtained by calculating the $k$ dimensional eigen-structure, 
e.g., using Singular Value Decomposition (SVD). Several results have been 
recently proposed for obtaining a low rank approximation 
of the data for summarization more efficiently than SVD.
However, very few of these algorithms are designed for 
a streaming setting~\cite{RelativeErrors:Ghashami:14,
LowRank:ClarksonWoodruf:09, feldman2013turning, 
SimpleDetMatSke:Liberty:13}.
}

Motivated by~\cite{feldman2013turning}, we adopt 
a tree formulation to approximate
the low rank subspace which best represents the data.   
Unlike other methods, the tree formulation has sufficiently 
low time complexity for a real-time learner, considering the constants. 
Using this formulation, the time complexity for the summarization 
and learning steps is constant (on average) in between observations. 
In the next section we formalize and prove time
and space guarantees for this approach.

\subsection{Coreset Tree}

The key idea here is that a low rank approximation 
can be ``approximated'' using a merge-and-reduce 
technique. The origin of this approach goes back to Bentley and
Saxe~\cite{compressable:saxe:80}, and was further adopted for 
coresets in~\cite{agarwal2004approximating}.
In this work, following Feldman et al.~\cite{feldman2013turning}, 
we generate a tree structure, which holds a summary of the 
streaming data.

\begin{figure}[t]
\begin{center}
\includegraphics[width=1\linewidth]{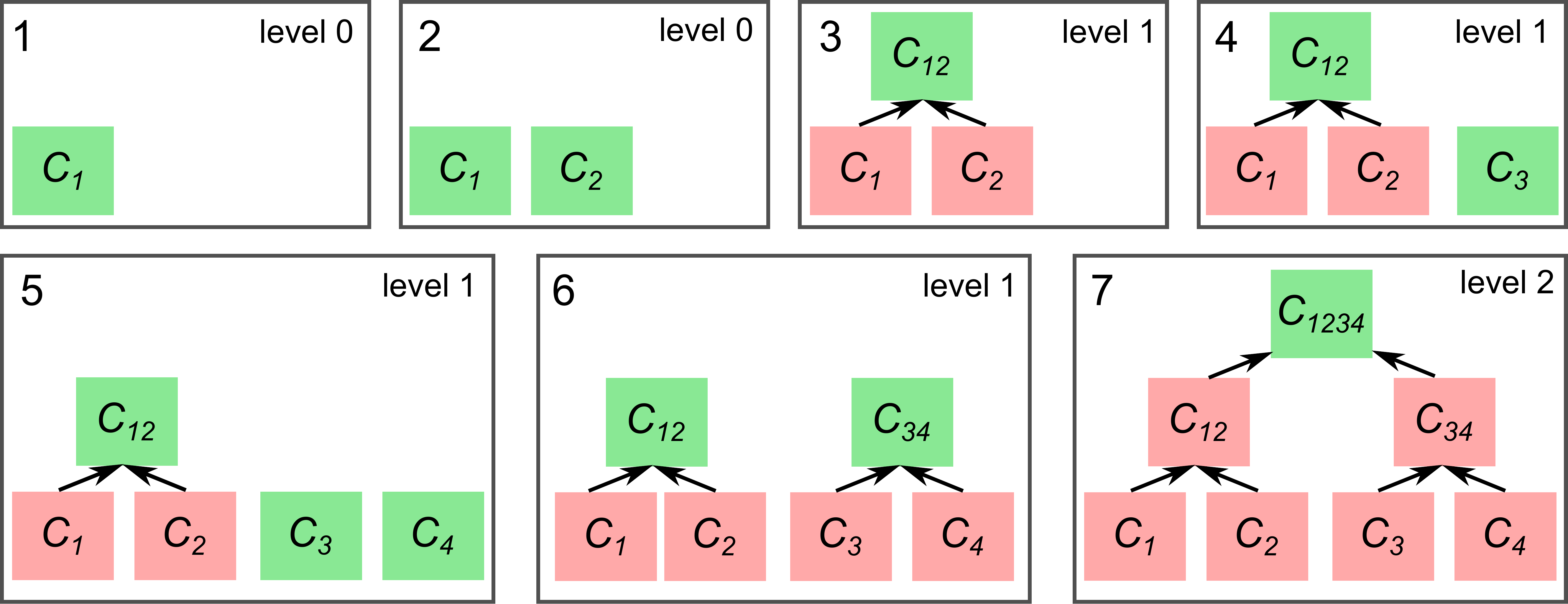}
\end{center}
\caption{\textbf{Tree construction:} Coresets are constructed using a  
merge-and-reduce approach. Whenever two coreset nodes are present 
at the same level, they are merged to construct a higher 
level coreset (shown in green). The nodes that have been merged 
are deleted from the memory (shown in red). We prove in 
Section~\ref{sec:theory} that the space requirement for
tree construction is $\mathcal{O}(\log p(t))$ in the worst
case, where $p(t)$ is
the number of data points at time $t$,
and on average the time complexity is constant per time step. }
\label{fig:coreset}
\end{figure}

We denote $C_i\in \mathbb{R}^{n\times d}$ as a leaf in 
the tree constructed from $n$ rows of $d$ dimensional features. 
The total number of data points captured until time $t$ is marked by
$p(t)$. For ease of calculations,  
we define $2{^{q(t)}}=\frac{p(t)}{n}$. Thus, $\lfloor q(t) \rfloor$ is the 
depth of the tree in time $t$. 
We construct a binary tree from the streaming data, and compress 
the data by merging the leaves, whenever two adjacent leaves get 
formed at the same level. 
If $p(t)=n2^{q_0}$, only the root 
node remains in memory after repeated merging of leaves.  
For the next $n2^{q_0}$ streamed points, 
the right side of the binary tree keeps growing and merging, until 
it is reduced to a single node again, when $n2^{q_0+1}$ data points 
arrive (Figure \ref{fig:coreset}).

If $C_i$ and $C_j$ are two $(1+\epsilon)$ coresets then their 
concatenation $\left[\frac{C_i}{C_j}\right]$ is 
also an $(1+\epsilon)$ coreset.
Since
\begin{eqnarray}
\left \Vert \left[\frac{C_i}{C_j} \right] Y \right \Vert_F^2 = \Vert C_iY \Vert_F^2 + \Vert C_jY \Vert_F^2,
\end{eqnarray} 
the proof follows trivially from the definition of a coreset.
This identity implies 
that two nodes which are $(1+\epsilon)$ approximation 
can be concatenated and then re-compressed 
back to size $n$.  
This is a key feature of the tree formulation, which repeatedly 
performs compression from $2n\times d$ to $n\times d$. Each   
computation can be considered constant in time and space.
However, this formulation also reduces the 
quality of the approximation, since 
the merged coreset is an $(1+\epsilon)^2$ 
approximation of the two nodes. For every level of the tree, 
a multiplication error of $(1+\epsilon)$ is introduced 
from the leaf node to the root.

\subsubsection{Data Approximation in the Tree} 
We denote the entire set up to time $t$ by $\cup_t C_i$, 
where $C_i$ are the leaves. We denote 
the coreset tree by $2 ^ {\cup_t C_i}$.
It follows that
\begin{Thm}
If $\forall i$ $\tilde C_i$ is a $(1+\epsilon/q(t))$ approximation of
$C_i$ then the collapse of the tree $2 ^ {\cup_t C_i}$ 
(i.e. the root) is a  $(1+\epsilon/3)$ coreset of $\cup_t C_i$  
\end{Thm}
\begin{proof}
We concatenate and compress sets of two leaves at 
every level of the tree. 
Compressing a $2n \times d$ 
points into $n \times d$ points introduces an   
additional multiplicative approximation of $(1+\epsilon/q(t))$.
Since the maximum depth of the tree at time $t$ is $q(t)$, and
\begin{eqnarray}
 \left( 1+ \frac{\epsilon}{q(t)}\right )^{q(t)} \le \exp(\epsilon /6) < (1 + \epsilon/3),
\end{eqnarray}
for $\epsilon<0.1$,
we conclude that if we compress $2n \times d$ points 
to $n \times d$ with an $(1+\epsilon/q(t))$ approximation, 
the root of the tree will be
a $(1+\epsilon/3)$ approximation of the entire data $\cup_t C_i$.
\end{proof}

\subsubsection{Space and Time Complexity} 
We now provide 
new proofs for the time and space complexity of 
coreset tree generation per time step and for the entire set.

\begin{Thm}
The space complexity to hold a  $2 ^ {\cup_t C_i}$ tree in memory is 
$\mathcal{O}(ndq(t))$, where $n$ is the
size of the coreset of points with dimension $d$.
\end{Thm}
\begin{proof}
In a coreset tree, two children of the same parent node 
cannot be in the memory stack simultaneously, which implies that, 
between two full collapses of the tree, the memory stack 
is increased at most by 
$\mathcal{O}(nd)$, which is the size of one leaf.
Hence, the space complexity $S(t)$ can be derived as
follows --
\begin{eqnarray}
 & & S(n2^{q(t)}) =1 \nonumber \\
 & & S(p(t))\le nd + \max_{\hat p(t)} S(\hat p(t))   \\
& & \hspace{0 cm}   n2^{q(t)-1}<\hat p(t)<n2^{q(t)}\,\,\, \& \,\,\,\ n2^{q(t)}< p(t) < n2^{q(t)+1}\nonumber
\end{eqnarray}
This recursive formulation results in a space requirement of
$S(p) \le \mathcal{O}(nd  \log(p(t)/n)) = \mathcal{O}(ndq(t))$. 
Assuming $n$ and $d$ are constants chosen
a priori, the complexity of the memory is 
logarithmic in the number of points seen so far.
\end{proof}

\newtheorem{Thm5}{Theorem 5}
\begin{Thm}
The time complexity of coreset tree construction between 
time step $t$ and $t+1$ is $\mathcal{O}( \log(p(t)/n) nd^2)$ in 
the worst case, 
which is logarithmic in $p$, the total number of points.
\end{Thm}
\begin{proof}
In the worst case, we perform $\log(p(t)/n)$ compressions 
from the lower level all the way to the root. The compressions 
are performed using SVD. 
Since the time complexity of each SVD algorithm is $\mathcal{O}(nd^2)$, assuming $n>d$, 
\begin{eqnarray}
T(t+1)-T(t) \le \mathcal{O}( \log(p(t)/n) nd^2) &=& \mathcal{O}(q(t)nd^2) \nonumber \\
&=& \mathcal{O}(q(t)),
\end{eqnarray}
where $n$ and $d$ are considered constants.
\end{proof}
\noindent
This scenario occurs for the last data point 
of a perfect subtree ($p(t)=n2^{q(t)}$).

Finally we wish to explore the total 
and average time complexity to summarize $p$ points. 
On average, the time complexity is constant and does not depend on $p$.
\begin{Thm}
The  time complexity for a \emph{complete collapse} of a tree to a
single coreset of size $n\times d$ is $\mathcal{O}(pd^2)$, where $d$ 
the dimension of each point and $p$ is the number 
of points. The average time complexity per point is constant.
\end{Thm}
\begin{proof}
$p$ points are split into $p/n$ leaves with $n \times d$ 
dimension each. The time complexity of SVD is $\mathcal{O}(nd^2)$, 
assuming $n>d$. A total of $p(t)/n$ SVDs will be computed 
for compressing a perfect tree. Hence the total time complexity 
$T(t)$ for all cycles until time $t$ becomes
\begin{eqnarray}
 T(t) = \mathcal{O}\left(\frac{p(t)}{n} nd^2\right) = \mathcal{O}(p(t)d^2)=\mathcal{O}(p(t))
\end{eqnarray}
assuming $d$ is constant. 
We further infer that the average time 
complexity $\hat T(t)$ is $\mathcal{O}(1)$ as
\begin{eqnarray}
\hat T(t) =  \frac{T(t)}{p(t)} = \mathcal{O}\left(\frac{p(t)d^2}{p(t)}\right)=\mathcal{O}(d^2)=\mathcal{O}(1).
\end{eqnarray}
\end{proof}
 
\section{CAT: Coreset Adaptive Tracker} 
\label{sec:implementation}
\begin{figure}[t]
\begin{center}
\includegraphics[width=1\columnwidth]{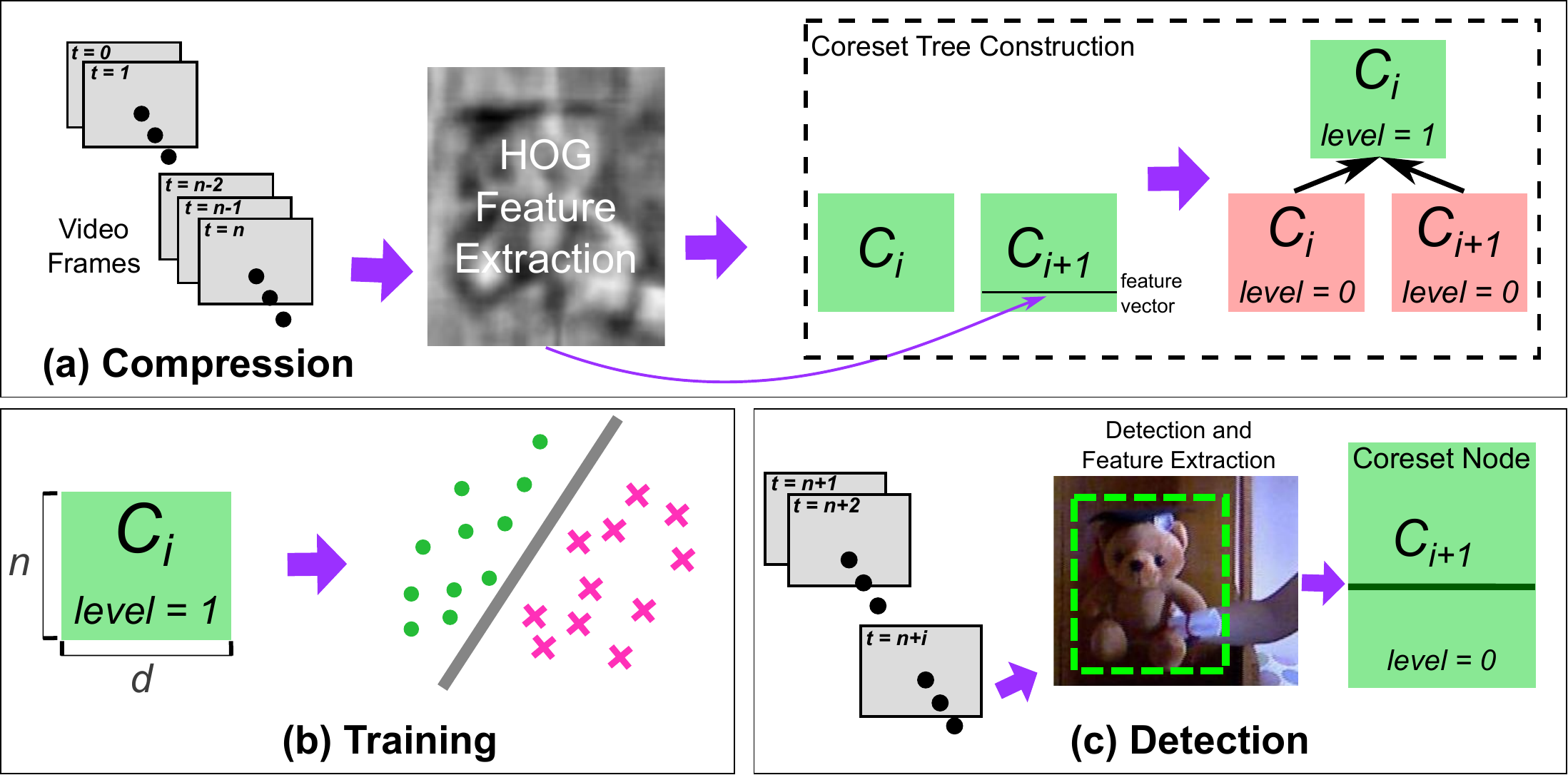}
\end{center}
\caption{
\textbf{Tracking by detection using coresets:} 
Our system for learning from streaming 
data contains three main components, which run in parallel. 
\textbf{(a) Compression using coresets:} For each frame, 
we extract HoG features from the object. 
After $n$ frames are processed, a new leaf is inserted into the coreset tree. 
The tree is compressed recursively by merging nodes at the 
same level using SVD. \textbf{(b) Training:} A linear SVM for object appearance 
is trained using data points from the coreset tree
after every $n$ frames are processed. The maximum size of the
training set is $2n\times d$. \textbf{(c) Detection:} The trained model is used for  
sliding window detection on every new frame. Features extracted 
from the detected area are inserted into the coreset tree.
}
\label{fig:system}
\end{figure}

In the previous section, we describe an efficient method to 
obtain a summarization of all the data till a point in a streaming
setting, with strong theoretical guarantees.  
In this section, we use the coreset tree for tracking by detection using a  
\emph{coreset appearance model} for the object. 

\subsection{Coreset Tree Implementation}
The coreset tree is stored as a stack of coreset nodes $S_C$. 
Each coreset node $S_C[i]$ 
has two attributes \emph{level} and \emph{features}. 
\emph{level} denotes the height of the coreset node in the tree, 
and \emph{features} are the image features extracted from a frame. 
In a streaming setting, once $n$ frames enter the frame stack, 
we create a coreset node with $level = 0$. We merge two coreset 
nodes when they have the same $level$. 

\subsection{Coreset Appearance Model}
\label{corset:appearance:model}
In the coreset tree, any node at level $q$
with size $n$ represents $n2^q$ frames, since 
it is formed after repeatedly merging $2^q$ leaf nodes. 
Nodes with different levels belong to different temporal
instances in streaming data. Nodes higher up in the tree 
have \emph{older} training samples, whereas a node 
with level 0 contains the last $n$ frames. This is not ideal for training a 
tracking system, since more recent frames should have a representation 
in the learning process. To account for recent data, we propose an hierarchical 
sampling of the entire tree, constructing an (at most) $2n \times d$-size summarization  
of the data seen until a given time. For each $S_C[i]$ we calculate 
$w_i = 2^{(\text{level}(S_C[i])-\text{level}(top(S_C))}$, and select 
the first $nw_i$ elements from $S_C[i]$. This summarized data 
constitutes our ``coreset appearance model'', which is used to 
train an object  detector.

\subsubsection{Advantages of Hierarchical Sampling} 
While the coreset tree formulation has been inspired by 
Feldman et al.~\cite{feldman2013turning}, our idea of  
hierarchical sampling of the entire tree allows to account
for recent data, which is crucial for many learning problems 
which need to avoid a temporal bias in the training set, and 
especially for tracking by detection. 
We demonstrate in Table~\ref{tab:compare} that hierarchical sampling 
improves the learning performance significantly as compared 
to just using the data from the root of the tree.

\subsection{System Implementation}
In the first frame , user input is used to select 
a region of interest (ROI) for the object to be tracked. 
We resize the ROI to a $64\times64$
square from which we extract $2\times2$ HOG features from 
the RGB channels. We also extract HoG features from small affine 
transformations of the input frame and add them to our frame 
stack. The affine transformations include scaling, translation 
and rotation. The number of affine transformations are determined 
by the number of additional frames 
required to make the frame stack size reach a maximum of 25\% of the 
total video length. For example, if the coreset size is 5\% of the total 
length of the video, we add 4 transformations per frame. This process 
gives us the advantage of having more---albeit synthetically generated---samples 
to train the classifier, and makes the tracker robust to 
small object scale changes and translations.

Initially, since no coreset appearance model has been trained, we need to use an 
alternate method for tracking (we use SCM~\cite{zhong2012scm} 
in our current implementation). 
Once first $n$ frames have been tracked, the first leaf of the
tree is formed. Data classification, compression, and training 
now run in parallel (Figure~\ref{fig:system}). 
These components are summarized below --

\begin{itemize}
\item \emph{Classification}: Detect the object in the 
stream from the \emph{last} trained classifier. 
 \item \emph{Compression}: Update a coreset tree, by inserting new
   leaves and summarizing data whenever possible.
 \item \emph{Training}: Train an SVM classifier from the 
hierarchical coreset appearance model.
 \end{itemize}

\begin{figure*}[t]
\begin{center}
\includegraphics[width=0.8\textwidth]{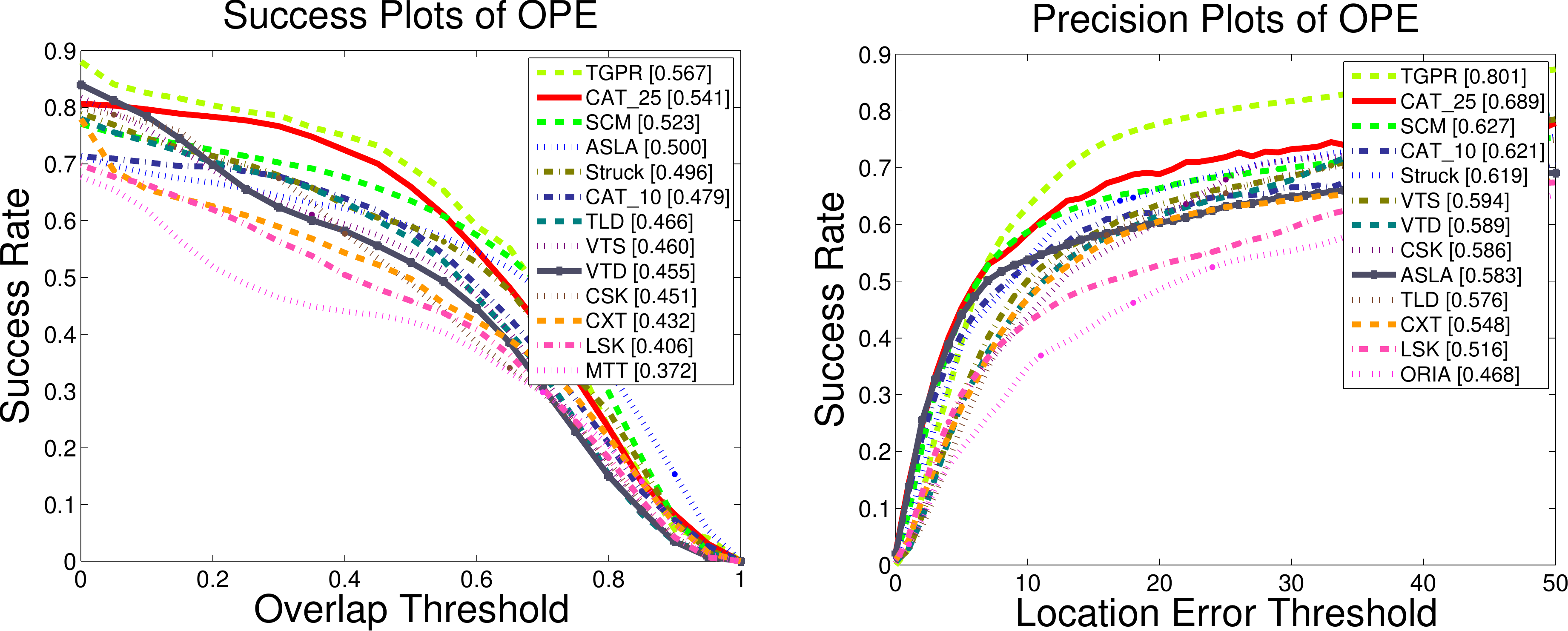}
\end{center}
\caption{Our methods (CAT\_10 and CAT\_25) obtains
competitive performance on the CVPR2013 Tracking Benchmark on videos with 500 frames or longer. 
CAT\_25 outperforms several top algorithms, such as Struck and SCM. 
}
\label{fig:cvpr13}
\end{figure*}

\begin{table*}[t]
\begin{center}
\begin{tabular}{|c|c|c|c|c|c|c|c|c|c|c|c|c|}
\hline
\multirow{2}{*} {\textbf{Algorithm}} & \textbf{Mean}
&\multicolumn{3}{|c|}{\textbf{Target Type}} &
\multicolumn{2}{|c|}{\textbf{Target Size}} &
\multicolumn{2}{|c|}{\textbf{Movement}} &
\multicolumn{2}{|c|}{\textbf{Occlusion}} &
\multicolumn{2}{|c|}{\textbf{Motion Type}} \\ \cline{3-13}
&\textbf{Score} &Human & Animal & Rigid &Large &Small &Slow &Fast &Yes
&No &Active &Passive \\ \hline
CAT\_25 &\textbf{0.52}& \textbf{0.52} & \textbf{0.55} & \textbf{0.63} & \textbf{0.49} & \textbf{0.49} & \textbf{0.65} & 0.38 & \textbf{0.41} & 0.58 & \textbf{0.59} & \textbf{0.44}\\
TGPR~\cite{gao2014tgpr} & 0.47 & 0.36 & 0.51 & 0.58 & 0.46 & \textbf{0.48} & 0.62 & \textbf{0.41}& 0.35 & \textbf{0.65} & 0.56 & 0.44\\
CAT\_10 &0.46& \textbf0.46 & 0.49 & \textbf 0.55 & \textbf 0.46 & 0.45 & 0.61 & 0.33 & 0.30 & 0.54 & 0.52 & 0.40\\
Struck~\cite{hare2011struck}& 0.44 & 0.35 &0.47& 0.53& 0.45& 0.44&
0.58& 0.39& 0.30 & 0.64& 0.54 &0.41\\
VTD~\cite{kwon2010visual}&0.45&0.31& 0.49 &0.54& 0.39& 0.46& 0.57&
0.37& 0.28& 0.63& 0.54& 0.39\\
RGB~\cite{song2013tracking}&0.42& 0.27 & 0.41 & 0.55 & 0.32 & 0.46 &
0.51& 0.36& 0.35& 0.47 & 0.56& 0.34 \\
TLD~\cite{kalal2012tracking}&0.38&0.29&0.35&0.44&0.32&0.38&0.52&0.30&
0.34&0.39&0.50&0.31\\
MIL~\cite{babenko2011robust}&0.37 & 0.32 & 0.37 & 0.38 & 0.37 & 0.35 & 0.46 & 0.31 & 0.26 &
0.49 & 0.40 & 0.34 \\
\hline
\end{tabular}
\end{center}
\caption{ 
We use the Princeton Tracking Benchmark~\cite{song2013tracking} 
to compare the precision of CAT with other algorithms using 
results reported by ~\cite{WinNT,liang2015adaptive}. 
We obtain state of the art results in overall precision 
and lead the benchmark in 9 out of 11 categories. 
} 
\label{lab:tab1}
\end{table*}

\subsubsection*{Classification}
If a classifier has been trained, 
we feed each sliding window to the classifier and obtain a  
confidence value. After hard thresholding and non-maximum 
suppression, we detect an ROI. To remove noise in the 
position of the ROI, we employ a Kalman filter.
We use Expectation Maximization (EM) to obtain its parameters
from the centers of the returned predictions 
of the previous $n$ frames. 
The EM returns learnt parameters for prediction, which are used to 
estimate the latent position of the 
bounding box center. This latent position is used 
to update the Kalman filter parameters for the next 
iteration. Once the center of the bounding box is obtained, 
we calculate the true ROI. We extract features from the ROI 
and feed it to a feature stack $S_P$. Until the first $n$ frames arrive
we obtain the ROI from  SCM~\cite{zhong2012scm} since no classifier has 
been trained. This process continues till the last frame.

\subsubsection*{Compression}
If $|S_P|>n$, we pop the top $n$ elements 
$S_{P,n}$ from the feature stack 
and push a coreset node containing elements 
$S_{P,n}$ into the coreset stack $S_C$ with $level =  0$.
If $level(S_C[i]) = level(S_C[i+1])$, they are merged by 
constructing a coreset from $S_C[i]$ and $S_C[i+1]$. We store 
the coreset in $S_C[i]$, and delete $S_C[i+1]$. 
The $level$ for node $S_C[i]$ is increased by one. This process 
continues recursively. 

\subsubsection*{Training}
If at least one  leaf 
has been constructed and the size of the coreset stack changes, 
we train a one-class linear SVM using samples from $S_C$. 
An (at most) $2n \times d$-size  appearance model is 
generated by hierarchical sampling of the tree as described in 
Section~\ref{corset:appearance:model}.

\section{Evaluation}
\label{sec:Evaluation}
We implement the tracking system in C++
with Boost, Eigen, OpenCV, pThreads, and libFreenect libraries.
{We evaluate the performance of CAT using three publicly available datasets 
at two coreset sizes---10\% and 25\%. We denote the tracker with 
coreset size 10\% as CAT\_10, and denote the tracker with 
coreset size 25\% as CAT\_25.

\subsection{CVPR2013 Tracking Benchmark}
We first evaluate our algorithm using the CVPR2013 Tracking 
Benchmark~\cite{wu2013online} which includes 50 annotated 
sequences. We use two evaluation metrics to compare the 
performance of tracking algorithms. The first metric is the 
\textbf{precision plot}, which shows the percentage of frames 
whose estimated location is within the 
given threshold distance of the ground truth. As the 
representative \textbf{precision score} for each tracker, 
we use the score for a threshold of 20 pixels.  
The second metric is the \textbf{success plot}. 
Given the tracked bounding box
$r_t$ and the ground truth bounding box $r_a$, the overlap
score is defined as $S = \frac{|r_t \cap r_a|}
{| r_t \cup r_a |}$, where $\cap$ and $\cup$ represent the 
intersection and union of two regions and $|\cdot|$ denotes 
the number of pixels in the region. We report the 
\textbf{success rate} as the fraction of frames whose 
overlap $S$ is larger than some threshold, with the threshold 
varying from 0 to 1. The area under curve (AUC) of the success plot is 
used to compare different algorithms. 

The strength of our method lies in tracking objects from videos 
with several hundred frames or longer. So we compare a 
variety of trackers using videos that are 500 frames or longer 
(21 out of 50 videos) using the 
One-Pass Evaluation (OPE) method~\cite{wu2013online}. 
Figure~\ref{fig:cvpr13} shows 
that we are placed second on this competitive dataset for 
longer videos in both metrics. We outperform important recent methods
including SCM~\cite{zhong2012scm} and Struck~\cite{hare2011struck}.

\subsection{Princeton Tracking Benchmark}

The Princeton Tracking Benchmark~\cite{song2013tracking} 
is a collection of 100 manually annotated RGBD tracking videos, with 214 
frames on average. The videos contain significant variation in 
lighting, background, object size, orientation, pose, 
motion, and occlusion. The reported score is an average over all frames of precision,
which is the ratio of overlap between the ground truth and detected bounding 
box around the object. We report the performance of all algorithms 
(including CAT) for RGB data using Princeton Tracking Benchmark 
evaluation server~\cite{WinNT} and from the 
recent work by Liang et al~\cite{liang2015adaptive} to ensure 
consistency in evaluation. CAT\_25 obtains state of the art 
performance on this dataset, and CAT\_10 ranks third 
(Table~\ref{lab:tab1}). Note that we obtain reasonable 
precision in presence of occlusions, even though we have no 
explicit way for occlusion handling. Sample detections 
are shown in Figure~\ref{fig:images}.

\subsection{TLD Dataset}
The coreset formulation summarizes all the data 
it has seen so far. Therefore we expect the coreset appearance model 
to adapt to pose and lighting variations as it trains using 
additional data, and perform well on very long videos. 
To demonstrate its adaptive behavior, we use 
the TLD dataset~\cite{kalal2012tracking}, which contains ten videos 
with $2685$ frames of labeled video on average.
For all experiments with the TLD dataset 
(Tables~\ref{tab:tld},~\ref{tab:sampling},~\ref{tab:tldshort}, and \ref{tab:long})  
we report the \textbf{success rate} at a threshold of 0.5. 
Tables~\ref{tab:tld}, \ref{tab:tldshort}, and \ref{tab:long} show that CAT\_25 
significantly outperforms the state of the art, including methods 
like TGPR~\cite{gao2014tgpr}, which perform better than CAT on the  
CVPR2013 Tracking Benchmark. CAT\_10 is placed third, and it's success rate
is similar to that of TLD. These results support our intuition that 
the ability of CAT to summarize \textit{all data} provides a boost 
in performance over long videos.

\subsection{Object Instance Recognition}
Since the coreset appearance model learns adaptively from the 
entire data and does not depend on recency as most tracking methods, 
it can be used for object instance recognition as well. 
In a qualitative experiment (Figure~\ref{fig:instance}), 
we show that a pre-learnt model can be used to detect 
the same object under varying background, illumination, 
pose, and deformations.

\begin{table}[t]
\begin{center}
\begin{tabular}{|c|c|c|c|c|c|}
\hline
\multirow{2}{*}{\textbf{Algorithm}} &\multicolumn{5}{|c|}{\textbf{Success Rate}} \\ \cline{2-6}
&Average&  Motocross& Panda & VW & Car chase  \\ \hline
\#Frames&2685&2665&3000&8576&9928\\\hline
CAT\_25&\textbf{0.66}&0.85&\textbf{0.58}&\textbf{0.67}&\textbf{0.85}\\
TLD~\cite{kalal2012tracking}&0.57&\textbf{0.86}&0.25
&0.67&0.81\\
CAT\_10 & 0.56 & 0.77 & 0.49 & 0.62 & 0.79\\
Struck~\cite{hare2011struck}&0.40&0.57&0.21&0.54&0.31\\
TGPR~\cite{gao2014tgpr}&0.38&0.62&0.12&0.62&0.18\\
SCM~\cite{zhong2012scm}&0.36&0.50&0.13&0.47&0.14\\\hline
\end{tabular}
\end{center}
\caption{
CAT obtains state of the art performance on the TLD 
dataset~\cite{kalal2012tracking}. We report 
the average success rate over all videos, as well as success rate
for the four longest videos.}
\label{tab:tld}
\end{table}

\begin{figure}
\begin{center}
\includegraphics[width=1\linewidth]{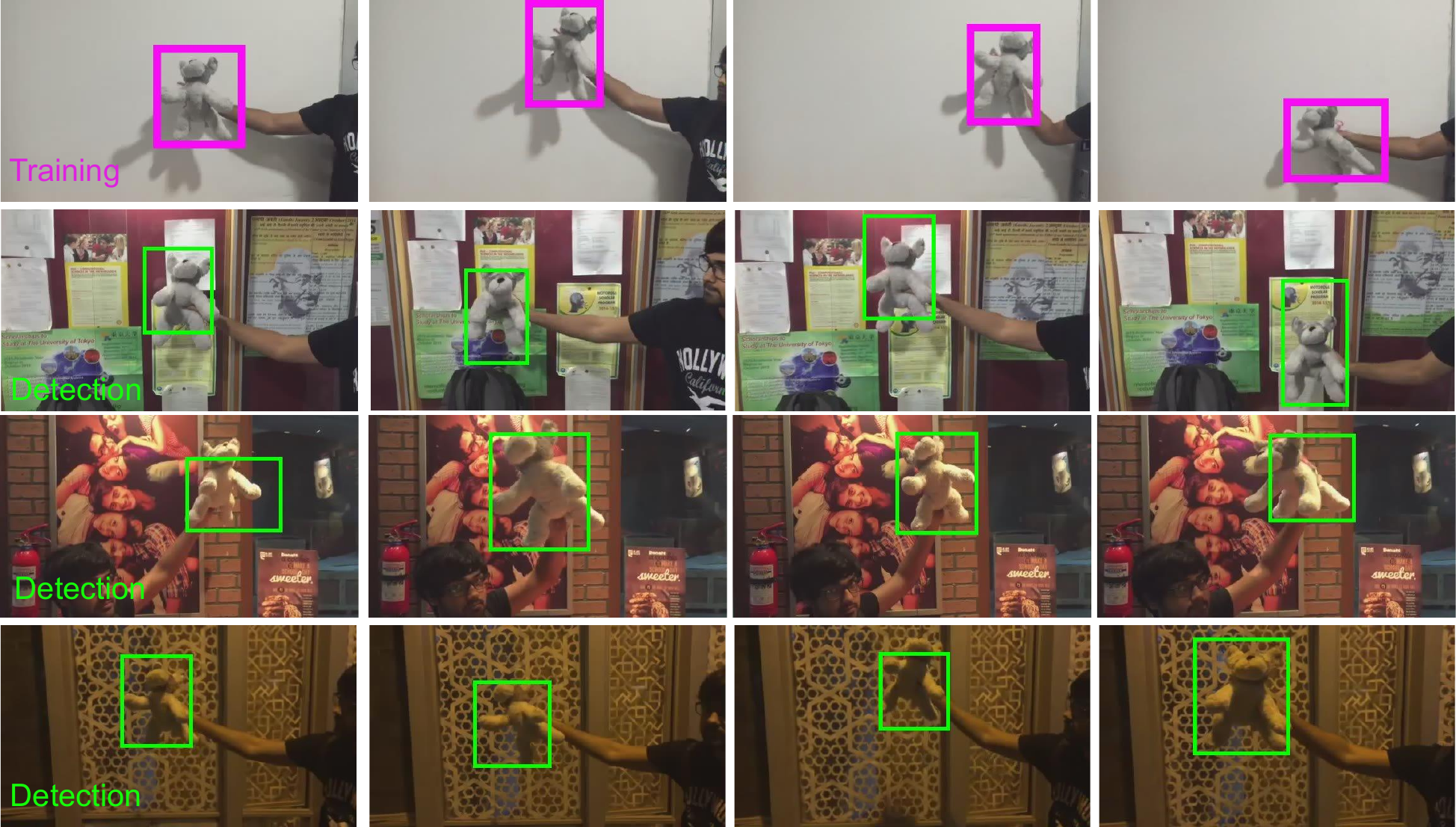}
\end{center}
\caption{\textbf{Object instance
  detection:}  In a qualitative experiment, 
we learn the coreset appearance model by 
tracking an object with plain background in 
$76$ frames of video (first row). 
This model can accurately detect and track 
the same object in different backgrounds (rows 2, 3, 4) 
and handle illumination changes (rows 3,4).}
\label{fig:instance}
\end{figure}

\subsection{Empirical Evaluation of Time Complexity}
The coreset formulation is used to train a learning algorithm 
in constant time and space (Section~\ref{sec:theory}). We compare 
the time complexity of training with the coreset tree and training 
using all the data points (Figure~\ref{fig:time}-(left)). 
For a linear SVM implementation with linear 
time complexity~\cite{chang2011libsvm}, the learning 
time for training with all data points can become prohibitively large,
in contrast to the constant, small time needed for learning from 
coresets. We pay for the reduced cost of 
training the learning algorithm by 
the additional time and space requirements of coreset tree 
formation. We prove in Section~\ref{sec:theory} that the time 
complexity of tree construction is constant on average. We validate 
this result empirically in Figure~\ref{fig:time}-(right).
Other coreset formulations~\cite{RelativeErrors:Ghashami:14}
may be used to reduce the space 
requirements to constant, at the cost of a higher value of the 
constant for time complexity. 

\subsection{Comparison with Sampling Methods}
In comparison to CAT, sampling can achieve reduction 
in space and time complexity without the additional 
overhead of coreset computation. We validate CAT versus 
random sampling and sub-sampling using the TLD dataset using 
multiple values of $n$, the coreset size. 
For random sampling (RS), we use $n$ random 
frames from all the frames seen so far.
For sub-sampling (SS), we use $n$ uniformly spaced frames. 
CAT consistently outperform both RS and SS (Table~\ref{tab:sampling}).
Thus, the summarization of data by coresets allows us to learn a 
more complex object appearance model as compared to sampling.  

\subsection{Coreset Size Selection}
The coreset size $n$ determines the time and space complexity 
of CAT. In this paper, we evaluate the 
performance of CAT at coreset sizes which are 10\% (CAT\_10) 
or 25\% (CAT\_25) of the total video size. Using such large values for 
coreset size is not a strict requirement, but 
rather an artifact of the datasets used for evaluation.
A majority of videos in the three standard datasets are 
short---114 out of 155 videos have less than 400 frames.   
We model the object using dense HOG features, which leads 
to a very high dimensional representation. Hence, the 
SVM requires sufficient number of data points for good 
prediction on such short videos. We set $n$ to 10\% or 25\% 
of the video size to obtain consistent performance across 
videos of various sizes. 

However, CAT maintains high detection accuracy with small  
coreset sizes as well. In Table~\ref{tab:tldshort}, we evaluate 
CAT with various coreset sizes---1\%, 2.5\%, 5\% and 
10\%---on the TLD dataset. Increasing the coreset size results in an 
increase in success rate for tracking, since the
model captures more information about object appearance. However, the tracking
performance is still satisfactory for coreset size as small as 1\% of the
video size, which is equal to 27 frames on average for the TLD dataset. 

Furthermore, CAT performs well, and in fact 
improves with time, on sufficiently long videos with 
small coreset sizes that are independent of the number of 
frames in a video. In Table~\ref{tab:long},
we compare the performance of CAT with fixed, small coreset size 
with two top performing methods on extremely long videos. We generate 
the extremely long videos from the four largest TLD
videos by appending clips that are 10\% in length of the original video. The
clips are randomly flipped, time-reversed, shifted in contrast
and brightness, translated by small amounts, and corrupted with Gaussian noise. 
CAT shows an increase in
accuracy with increasing number of frames. In contrast, the success rate
of TLD and TGPR reduces with an increase in number of frames. The
maximum coreset size in the 100K experiment is 2.5\% of the
original video. Yet CAT maintains excellent tracking accuracy. 

In summary, the impressive performance of CAT---a simple tracker based on 
appearance models trained with small coreset sizes---in these two 
experiments demonstrates the utility of the coreset formulation.  

\begin{figure}
\begin{center}
\includegraphics[width=1\linewidth]
{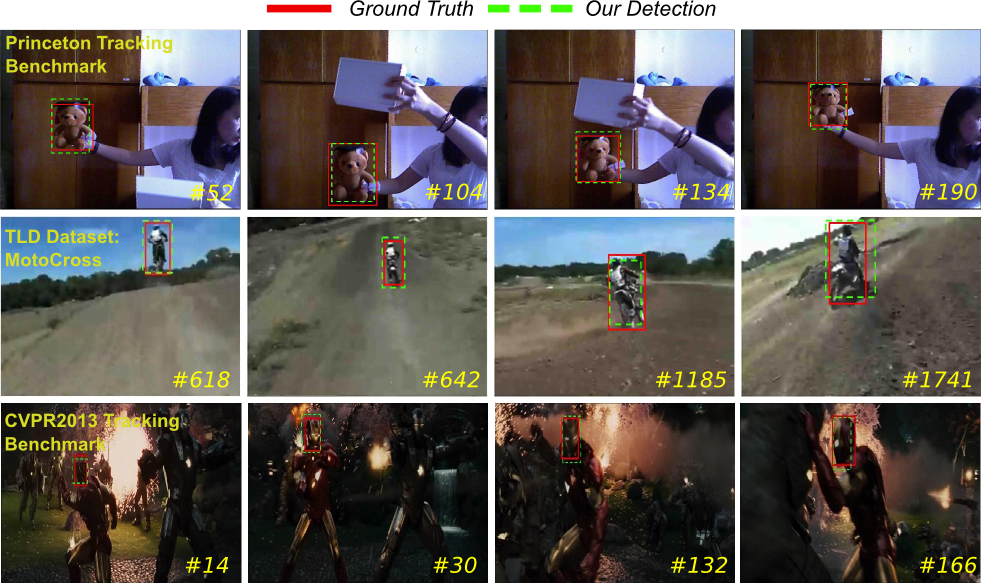}
\end{center}
\caption{\textbf{Sample detections:} 
We demonstrate excellent performance on the 
CVPR2013 Tracking Benchmark, the Princeton Tracking Benchmark 
and the TLD dataset.}
\label{fig:images}
\end{figure}

\begin{table}[t]
\begin{center}
\begin{tabular}{| m{3.5cm} | c | c |}
\hline
{\textbf{Algorithm}} & 
{\textbf{Success Rate}} & 
{\textbf{Precision}} 
\\ \hline
Learning from root of  
the tree~\cite{feldman2013turning} &  0.398 & 0.434   \\\hline
Learning with hierarchical sampling (CAT)  & 0.524 & 0.677\\
\hline
\end{tabular}
\end{center}
\caption{Our hierarchical sampling method allows for more 
representation to recent data in the learning process. 
In the original formulation~\cite{feldman2013turning}, 
the classifier would be trained from the root of 
the coreset tree alone, ignoring more recent data. We 
demonstrate that our method    
provides better accuracy as compared to the original
formulation using the CVPR2013 Tracking Benchmark. 
}
\label{tab:compare}
\end{table}

\begin{table}[t]
\begin{center}
\begin{tabular}{|c|c|c|c|c|c|}
\hline
\multirow{2}{*}{\textbf{Algorithm}} &\multicolumn{5}{|c|}{\textbf{\%
Frames Used For Training}} \\ \cline{2-6}
&25 & 37.5 & 50 & 62.5 & 75 \\ \hline

\multicolumn{6}{|c|}{\textit{Coreset Size =512}} \\ \hline
CAT &0.408&0.449&0.483&0.534&0.582\\
SS&0.332&0.358&0.396&0.408&0.424\\
RS&0.218&0.221&0.251&0.279&0.301\\ \hline
\multicolumn{6}{|c|}{\textit{Coreset Size = 768}} \\ \hline
CAT &0.502&0.559&0.604&0.637&0.668\\
SS&0.378&0.415&0.438&0.474&0.502\\
RS&0.257&0.299&0.326&0.368&0.385\\ \hline
\multicolumn{6}{|c|}{\textit{Coreset Size = 1024}} \\ \hline
CAT &0.546&0.621&0.657&0.701&0.732\\
SS&0.418&0.492&0.54&0.587&0.633\\
RS&0.321&0.36&0.381&0.419&0.458\\ \hline
\end{tabular}
\end{center}
\caption{We show that CAT consistently obtains better success rate than 
random sampling (RS) and sub-sampling (SS) using the TLD dataset.
CAT also shows a greater marginal increase in success rate with
increase in $n$ as compared to RS and SS, since it generates a
better summarization of the data.}
\label{tab:sampling}
\end{table}

\begin{figure}
\begin{center}
\includegraphics[width=1.0\columnwidth]{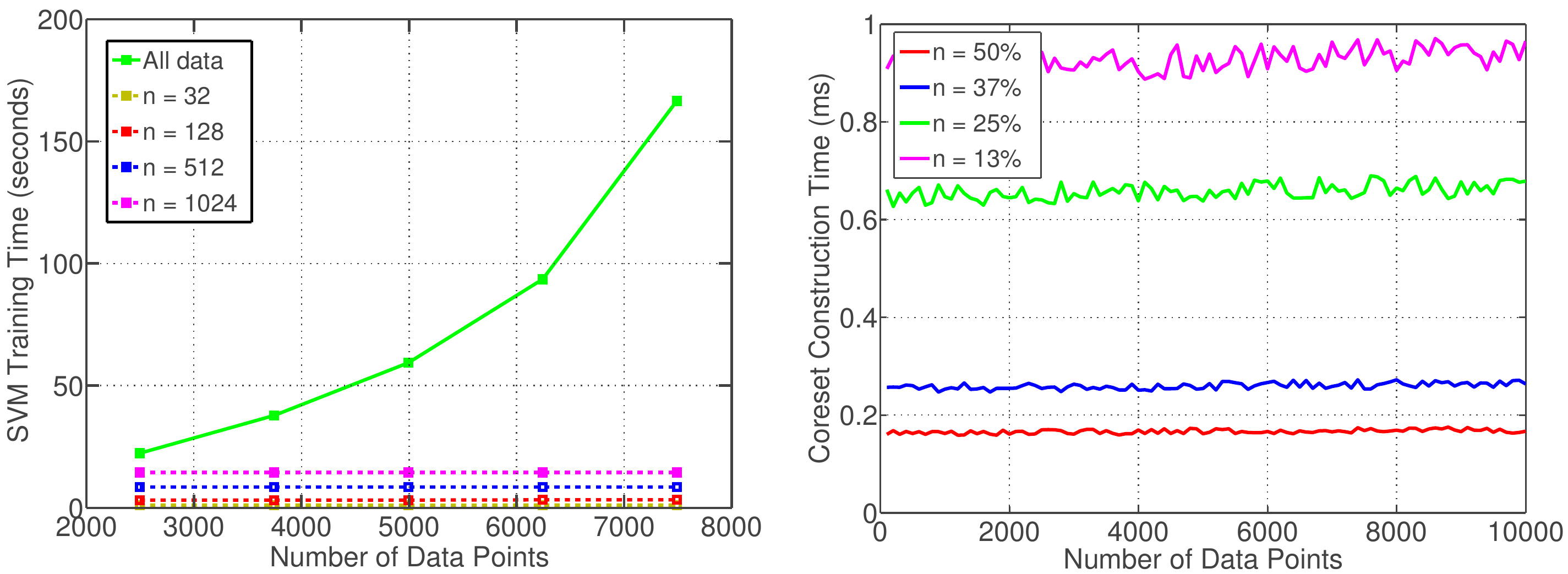}
\end{center}
   \caption{We validate the theoretical bounds for 
time complexity with empirical analysis using average 
over all videos from TLD dataset.
 (Left) The coreset tree allows us to
   train an SVM in constant time (depending on the coreset size $n$),
 while the learning time becomes 
prohibitively large while training with all data points. 
(Right) The time complexity of coreset construction is constant on
average. We demonstrate this at various coreset sizes $n$,  
expressed as percentage of the total video size, and averaged over 
all videos in the TLD dataset. 
}
\label{fig:time}
\end{figure}

\begin{table}[t]
\centering
\begin{tabular}{|l|c|c|c|c|c|}
\hline
\multirow{2}{*}{\textbf{Coreset Size}} & \multicolumn{5}{|c|}{\textbf{Success Rate}} \\ \cline{2-6}
& Mean   & Motocross   & VW & Panda   & Carchase  \\ \hline
1\%  & 0.49 & 0.74 & 0.48            & 0.59 & 0.73 \\
2.5\% & 0.51 & 0.76 & 0.49            & 0.59 & 0.74 \\
5\% & 0.53 & 0.77 & 0.49            & 0.61 & 0.74 \\
10\% (CAT\_10) & 0.56 & 0.77 & 0.49            & 0.62 & 0.79 \\
25\% (CAT\_25) &\textbf{0.66}&0.85&\textbf{0.58}&\textbf{0.67}&\textbf{0.85}\\
\hline
\end{tabular}
\caption{{We evaluate CAT at several coreset sizes as the percent of video 
size using the TLD dataset. CAT performs satisfactorily at coreset size as small
as 1\% of the video size.}}
\label{tab:tldshort}
\end{table}

\begin{table}[t]
\centering
\begin{tabular}{|c|l|l|l|l|l|}
\hline
\multirow{2}{*}{\textbf{Algorithm}} & \multicolumn{5}{|c|}{\textbf{Number of Frames}} \\ \cline{2-6}
& 10K   & 20K   & 50K & 75K   & 100K  \\ \hline
CAT (\textit{n}=500)  & 0.522 & 0.559 & 0.578            & 0.586 & 0.594 \\
CAT (\textit{n}=1000) & 0.606 & 0.634 & 0.658            & 0.669 & 0.673 \\
CAT (\textit{n}=1500) & 0.627 & 0.674 & 0.697            & 0.701 & 0.702 \\
CAT (\textit{n}=2000) & 0.685 & 0.702 & 0.718            & 0.725 & 0.726 \\
CAT (\textit{n}=2500) & 0.713 & 0.724 & 0.736            & 0.743 & 0.744 \\
TLD~\cite{kalal2012tracking}& 0.579 & 0.573 & 0.569            & 0.567 & 0.567 \\
TGPR~\cite{gao2014tgpr}& 0.382 & 0.368 & 0.365            & 0.364 & 0.362 \\ \hline
\end{tabular}
\caption{Success Rate Comparison: CAT performs well, and in fact improves with time, on very long  
videos with small, fixed coreset size which is independent of 
the total number of frames.}
\label{tab:long}
\end{table}

\section{Discussion}

\textbf{Tracking in Long Videos:} A majority of 
tracking algorithms are unable to directly make use 
of data from all the previous frames. 
At best, they retain a few `good' frames to 
learn~\cite{supancic2013self} or maintain an auxiliary set 
of earlier frames~\cite{gao2014tgpr}. Our method can retain a 
summary of all frames seen so far using logarithmic space with 
an efficient algorithm. 
This is an important advantage for  
practical systems with space and time constraints.

\textbf{Simplicity:} The utility of 
the coreset formulation is evident from the fact that we 
obtain state of the art experimental results 
with a simple tracker that consists of a linear classifier and 
a Kalman filter. The primary goal of this work is 
to explore the benefit of coresets for learning from 
streaming visual data, rather than to compete 
on tracking benchmarks. Additional improvement could 
be obtained via choosing the optimal coreset size or exploring 
various hierarchical sampling methods. Performance may also 
improve by incorporating traditional methods for handling 
occlusion and drift.

\textbf{Real-time Performance:} 
Our current implementation, running on 
commodity hardware (Intel i7 Processor, 8 GB RAM), 
performs coreset construction and 
SVM training in parallel at $\approximate$20 FPS for 
small coreset sizes ($n \leq 64$). 
Due to the high computational complexity of 
sliding window detection, the total speed is 
$\approximate$3 FPS. Faster detection methods (e.g., 
efficient subwindow search~\cite{lampert2008beyond} and   
parallelization~\cite{wojek2008sliding}) can 
be easily incorporated into our system 
to achieve real-time performance. 

\section{Concluding Remarks}
In this paper, we proposed an efficient method for 
training object appearance models using streaming data. The key
idea is to use coresets to compress object features through time to
a constant size and perform learning in constant time using 
hierarchical sampling of coresets. Our method is ideal for efficient 
tracking in long videos in real world scenarios.  
Since we maintain a compact summary of all previous data, our 
learning model improves with time. We provide experimental 
validation for this insight with strong performance on 
the TLD dataset which contains videos with $2685$ frames on average.
The ideas described in this paper complement 
the current expertise in the computer vision community on developing 
trackers and detectors.
Combining our framework with existing methods should 
lead to fruitful online approaches, particularly when `old' data 
should be taken into account. While we explore the use of 
coresets for object tracking in this paper, we hope that 
our technique can be adapted for a variety of applications, such as image 
collection summarization, video summarization and more.

{\normalsize  
\bibliographystyle{ieee}
\bibliography{tracking,coreset}
}

\end{document}